\documentclass[runningheads]{llncs}
\usepackage[utf8]{inputenc}

\usepackage{cite}
\usepackage{amsmath}
\usepackage{mathtools}
\usepackage{subcaption}
\usepackage{graphicx}
\usepackage{hyperref}

\usepackage{color,soul}
\usepackage[table]{xcolor}

\begin{document}

\title{Deep Active Inference for Partially Observable MDPs\thanks{1st International Workshop on Active inference, European Conference on Machine Learning (ECML/PCKDD 2020)}}


\author{Otto van der Himst \and
Pablo Lanillos}

\authorrunning{O. van der Himst, P. Lanillos}

\institute{
Department of Artificial Intelligence\\
Donders Insitute for Brain, Cognition and Behaviour\\
Radboud University\\
Montessorilaan 3, 6525 HR Nijmegen, the Netherlands\\
\email{o.vanderhimst@student.ru.nl} \\
\email{p.lanillos@donders.ru.nl}}


\maketitle              

\begin{abstract}
Deep active inference has been proposed as a scalable approach to perception and action that deals with large policy and state spaces. However, current models are limited to fully observable domains. In this paper, we describe a deep active inference model that can learn successful policies directly from high-dimensional sensory inputs. The deep learning architecture optimizes a variant of the expected free energy and encodes the continuous state representation by means of a variational autoencoder. We show, in the OpenAI benchmark, that our approach has comparable or better performance than deep Q-learning, a state-of-the-art deep reinforcement learning algorithm.

\end{abstract}

\section{Introduction}
Deep active inference (dAIF) \cite{Ueltzh_ffer_2018, sancaktar2019endtoend, MILLIDGE2020102348, alex2019scaling, catal2020deep, fountas2020deep} has been proposed as an alternative to Deep Reinforcement Learning (RL) \cite{mnih2013playing, 8103164} as a general scalable approach to perception, learning and action. The active inference mathematical framework, originally proposed by Friston in \cite{Friston2010AaB}, relies on the assumption that an agent will perceive and act in an environment such as to minimize its free energy \cite{Friston2010TFEP}. Under this perspective, action is a consequence of top-down proprioceptive predictions coming from higher cortical levels, i.e., motor reflexes minimize prediction errors \cite{Adams2012PnC}.

On the one hand, works on dAIF, such as  \cite{sancaktar2019endtoend, lanillos2020robot, rood2020deep}, have focused on scaling the optimization of the Variational Free-Energy bound (VFE), as described in \cite{Friston2010AaB, oliver2019active}, to high-dimensional inputs such as images, modelling the generative process with deep learning architectures. This type of approach preserves the optimization framework (i.e., dynamic expectation maximization \cite{PMID:18434205}) under the Laplace approximation by exploiting the forward and backward passes of the neural network. Alternatively, pure end-to-end solutions to VFE optimization can be achieved by approximating all the probability density functions with neural networks \cite{Ueltzh_ffer_2018, MILLIDGE2020102348}.

On the other hand, Expected Free Energy (EFE) and Generalized Free Energy (GFE) were proposed to extend the one-step ahead implicit action computation into an explicit policy formulation, where the agent is able to compute the best action taking into account a time horizon \cite{Parr2019GFEaAI}. Initial agent implementations of these approaches needed the enumeration over every possible policy projected forward in time up to the time horizon, resulting in significant scaling limitations. As a solution, deep neural networks were also proposed to approximate the densities comprising the agent’s generative model \cite{Ueltzh_ffer_2018, sancaktar2019endtoend, MILLIDGE2020102348, alex2019scaling, catal2020deep, fountas2020deep}, allowing active inference to be scaled up to larger and more complex tasks.

However, despite the general theoretical formulation, current state-of-the-art dAIF, has only been successfully tested in toy problems with fully observable state spaces (Markov Decision Processes, MDP). Conversely, Deep Q-learning (DQN) approaches \cite{mnih2013playing} can deal with high-dimensional inputs such as images.

Here, we propose a dAIF model\footnote{The code is available on: \url{https://github.com/Grottoh/Deep-Active-Inference-for-Partially-Observable-MDPs}} that extends the formulation presented in \cite{MILLIDGE2020102348} to tackle problems where the state is not observable\footnote{We formulate image-based estimation and control as a POMDP---See \cite{SDMIA15-Hausknecht} for a discussion}  (i.e. Partially Observable Markov Decision Processes, POMDP), in particular, the environment state has to be inferred directly high-dimensional from visual input. The agent’s objective is to minimize its EFE into the future up to some time horizon T similarly as a receding horizon controller. We compared the performance of our proposed dAIF algorithm in the OpenAI CartPole-v1 environment against DQN. We show that the proposed approach has comparable or better performance depending on observability.

\section{Deep Active Inference Model} \label{free-energy-model}
\begin{figure}[hbtp!]
\includegraphics[width=\textwidth]{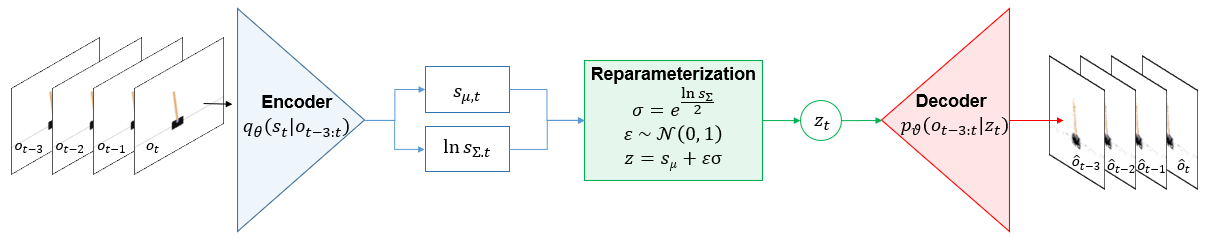}
\caption{Observations-state neural network architecture. The VAE encodes the visual features that are relevant to reconstruct the input images. The encoder network encodes observations to a state representation of the environment. The decoder reconstructs the input observations from this representation.} \label{fig:vae_architecture}
\end{figure}

\noindent We define the active inference agent’s objective as optimizing its variational free energy (VFE) at time t, which can be expressed as:
\begin{align}
    -F_t =& D_{KL}[q(s,a)\|p(o_t,s_{0:t},a_{0:t})] \\
    =& -E_{q(s_t)}[\ln{p(o_t|s_t)]} + D_{KL}[q(s_t)\|p(s_t|s_{t-1},a_{t-1})] \nonumber\\
    &+  D_{KL}[q(a_t|s_t\|p(a_t|s_t)]
\label{eq2}
\end{align}

\noindent Where $o_t$ is the observation at time $t$, $s_t$ is the state of the environment, $a_t$ is the agent’s action and $E_{q(s_t)}$  is the expectation over the variational density $q(s_t)$.

We approximate the densities of Eq. \ref{eq2} with deep neural networks as proposed in \cite{Ueltzh_ffer_2018, MILLIDGE2020102348, alex2019scaling}. The first term, containing densities $q(s_t)$ and $p(o_t|s_t)$ concerns the mapping of observations to states, and vice-versa. We capture this objective with a variational autoencoder (VAE). A graphical representation of this part of the neural network architecture is depicted in Fig. \ref{fig:vae_architecture} – see the appendix for network details.

We can use an encoder network $q_\theta(s_t|o_{t-3:t})$ with parameters $\theta$ to model $q(s_t)$, and we can use a decoder network $p_\vartheta(o_{t-3:t}|z_t)$ with parameters $\vartheta$ to model $p(o_t|s_t)$. The encoder network encodes high-dimensional input as a distribution over low-dimensional latent states, returning the sufficient statistics of a multivariate Gaussian, i.e. the mean $s_\mu$ and variance $s_\Sigma$. The decoder network consequently reconstructs the original input from reparametrized sufficient statistics $z$. The distribution over latent states can be used as a model of the environment in case the true state of an environment is inaccessible to the agent (i.e. in a POMDP).

The second term of Eq. \ref{eq2} can be interpreted as state prediction error, which is expressed as the Kullback-Leibler (KL) divergence between the state derived at time t and the state that was predicted for time t at the previous time point. In order to compute this term the agent must, in addition to the already addressed $q(s_t)$, have a transition model $p(s_t |s_{t-1},a_{t-1})$, which is the probability of being in a state given the previous state and action.  We compute the MAP estimate with a feedforward network $\hat{s}_t = f_{\phi}(s_{\mu,t-1},a_{t-1})$. To compute the state prediction error, instead of using the KL-divergence over the densities, we use the Mean-Squared-Error (MSE) between the encoded mean state $s_{\mu}$ and the predicted state $\hat{s}$ returned by $f_{\phi}$

\begin{figure}
\includegraphics[width=\textwidth]{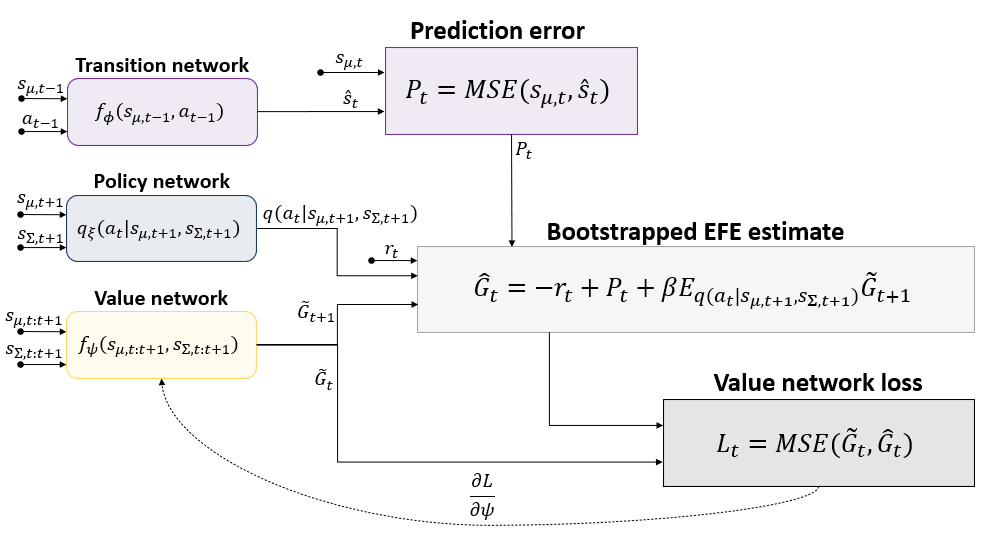}
\caption{Computing the gradient of the value network with the aid of a bootstrapped EFE estimate.} \label{fig:efe_computation}
\end{figure}

The third and final term contains the last two unaddressed densities $q(a_t|s_t)$ and $p(a_t|s_t)$. We model variational density $q(a_t|s_t)$ using a feedforward neural network $q_\xi(a_t|s_{\mu,t},s_\Sigma)$ parameterized by $\xi$, which returns a distribution over actions given a multivariate Gaussian over states. Finally, we model action conditioned by the state or policy $p(a_t|s_t)$. According to the active inference literature, if an agent that minimizes the free energy does not have the prior belief that it selects policies that minimize its (expected) free energy (EFE), it would infer policies that do not minimize its free energy \cite{Parr2019GFEaAI}. Therefore, we can assume that our agent expects to act as to minimize its EFE into the future. The EFE of a policy $\pi$ after time $t$ onwards can be expressed as:

\begin{equation}\label{eq3}
\begin{split}
    G_\pi &= \sum_{\tau>t}G_{\pi,t} \\
    G_{\pi,\tau} &= \underbrace{-\ln{p(o_\tau)}}_{-r_\tau} + D_{KL}[q(s_\tau|\pi)\|q(s_\tau|o_\tau)]
\end{split}
\end{equation}

\noindent Note that the EFE has been transformed into a RL instance by substituting the negative log-likelihood of an observation $-\ln{p(o_\tau)}$ (i.e. surprise) by the reward $r_\tau$ \cite{MILLIDGE2020102348, Frison2012AIaA}.
Since under this formulation minimizing one’s EFE involves computing one’s EFE for each possible policy $\pi$ for potentially infinite time points $\tau$, a tractable way to compute $G_\pi$ is required. Here we estimate $G_\pi$ via bootstrapping, as proposed in \cite{MILLIDGE2020102348}. To this end the agent is equipped with an EFE-value (feedforward) network $f_\psi(s_{\mu,t},s_{\Sigma,t})$ with parameters $\psi$, which returns an estimate $\Tilde{G}_t$ that specifies an estimated EFE for each possible action. This network is trained with the aid of a bootstrapped EFE estimate $\hat{G}_t$, which consists of the free energy for the current time step, and a $\beta\in(0,1]$  discounted value net approximation of the free energy expected under $q(a|s)$ for the next time step:

\begin{equation}\label{eq4}
    \hat{G}_t = -r_t + D_{KL}[q(s_t)\|q(s_t|o_t)] + \beta E_{q(a_{t+1}|s_{t+1})}\Tilde{G}_t
\end{equation}

\noindent In this form the parameters of $f_\psi(s_{\mu,t},s_{\Sigma,t})$ can be optimized through gradient descent on (see Fig. \ref{fig:efe_computation}):

\begin{equation}\label{eq5}
    L_t = MSE(\Tilde{G}_t,\hat{G}_t)
\end{equation}

The distribution over actions can then at last be modelled as a precision-weighted Boltzmann distribution over our EFEs estimate \cite{MILLIDGE2020102348, Parr2019GFEaAI}:

\begin{equation}
    p(a_t|s_t) = \sigma{(-\gamma \tilde{G}_t)}
\end{equation}

\begin{figure}[hbtp!]%
\includegraphics[width=\textwidth]{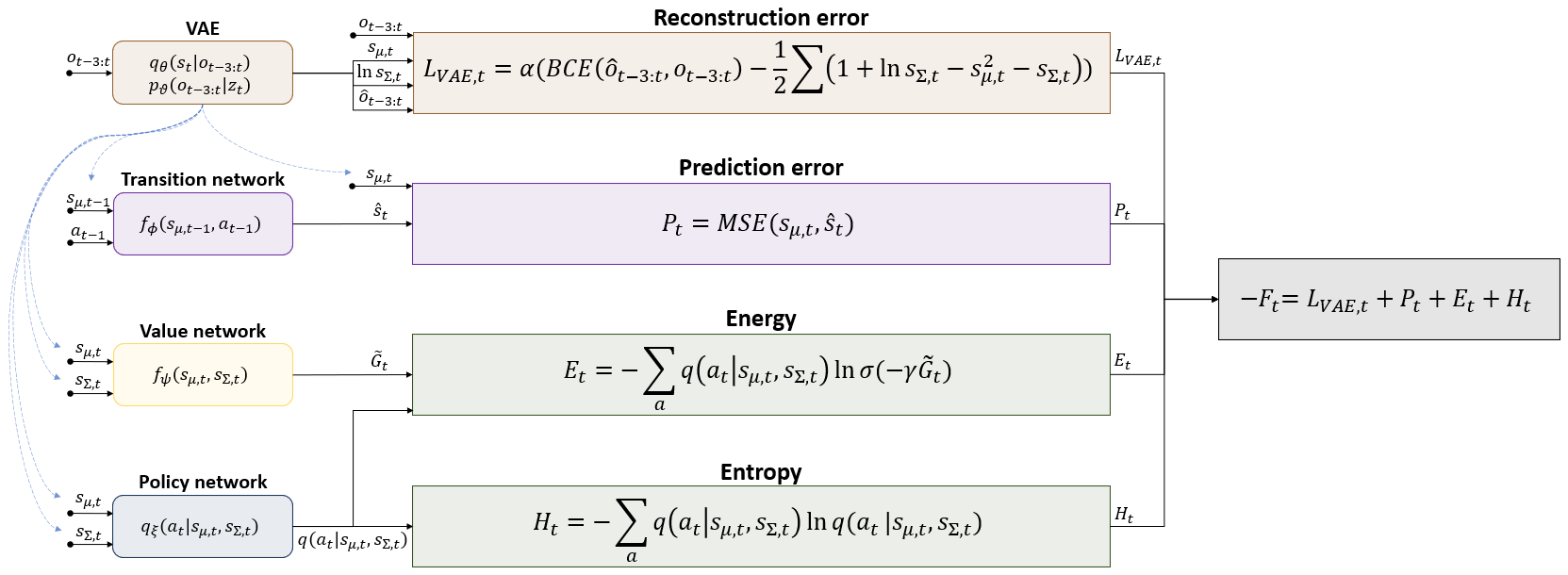}
\caption{Variational Free Energy computation using the approximated densities. The VAE encodes high-dimensional input as a latent state space, which is used as input to the other networks. Note that the third term of Eq. \ref{eq6} ($D_{KL}[q_\xi(a_t|s_{\mu,t},s_{\Sigma,t})\|\sigma(-\gamma f_\psi(s_{\mu,t},s_{\Sigma,t}))]$) has been split into an energy and an entropy term\protect\footnotemark.}
\label{fig:vfe_computation}
\end{figure}

\noindent Finally, Eq. \ref{eq2} is computed with the neural network density approximations as – See Fig. \ref{fig:vfe_computation}.
\begin{align}
    -F_t =& -E_{q_\theta(s_t|o_{t-3:t})}[\ln{p_\vartheta(o_{t-3:t}|z_t)}] \nonumber\\
    &+ MSE(s_{\mu,t},f_\phi(s_{\mu,t-1},a_{t-1})) \nonumber\\
    &+ D_{KL}[q_\xi(a_t|s_{\mu,t},s_{\Sigma,t})\|\sigma(-\gamma f_\psi(s_{\mu,t},s_{\Sigma,t}))]
\label{eq6}
\end{align}
\noindent Where $s_{\mu,t}$ and $s_{\Sigma,t}$ are encoded by $q_\theta(s_t|o_{t-3:t})$.

\footnotetext{Any KL divergence can be split into an energy term and an entropy term: 
\protect\begin{align}
    D_{KL}[P\|Q] &= \sum_{x\in X} P(x) \ln{\dfrac{P(x)}{Q(x)}} \nonumber \\
    &= \underbrace{-\sum_{x\in X} P(x) \ln{Q(x)}}_{Energy} \underbrace{- \sum_{x\in X} P(x) \ln{P(x)} }_{Entropy} \nonumber
\protect\end{align}
}

\section{Experimental Setup}
\begin{figure*}[hbtp!]
    \centering
    \begin{subfigure}{0.29\textwidth}
    \centering
        \includegraphics[width=1\linewidth]{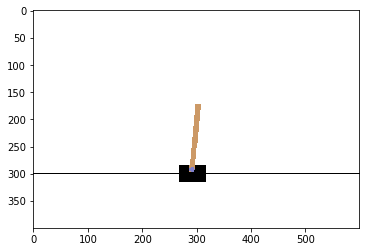}
    \end{subfigure}
    \begin{subfigure}{0.4\linewidth}
        \includegraphics[width=\linewidth]{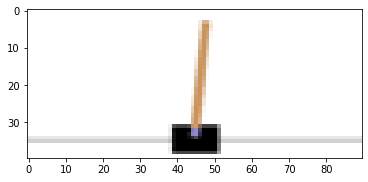}
    \end{subfigure}
    \caption{Cartpole-v1 benchmark (left) and cropped visual input used (right).}
    \label{fig:cartpole}
\end{figure*}

To evaluate the proposed algorithm we used the OpenAI Gym’s CartPole-v1, as depicted in Fig. \ref{fig:cartpole}. In the CartPole-v1 environment, a pole is attached to a cart that moves along a track. The pole is initially upright, and the agent’s objective is to keep the pole from tilting too far to one side or the other by increasing or decreasing the cart’s velocity. Additionally, the position of the cart must remain within certain bound. An episode of the task terminates when the agent fails either of these objectives, or when it has survived for 500 time steps. Each time step the agent receives a reward of 1.

The CartPole state consists of four continuous values: the cart position, the cart velocity, the pole angle and the velocity of the pole at its tip. Each run the state values are initialized at random within a small margin to ensure variability between runs. The agent can exact influence on the next state through two discrete actions, by pushing the cart to the left, or by pushing it to the right.

Tests were conducted in two scenarios: 1) an MDP scenario in which the agent has direct access to the state of the environment, and 2) a POMDP scenario in which the agent does not have direct access the environment state, and instead receives pixel value from which it must derive meaningful hidden states. By default, rendering the CartPole-v1 environment  returns a $3\times400\times600$ (color, height, width) array of pixel values. In our experiments we provide the POMDP agents with a downscaled and cropped image. There the agents receive a $3\times37\times85$ pixel value array in which the cart is centered until it comes near the left or right border.

\section{Results}

The performance of our dAIF agents was compared against DQN agents for the MDP and the POMDP scenarios, and against an agent that selects it actions at random. Each agent was equipped with a memory buffer and a target network \cite{Mnih2015HLCtDRL}. The memory buffer stores transitions from which the agent can sample random batches on which to perform batch gradient descent. The target network is a copy of the value network of which the weights are not updated directly through gradient descent, but are instead updated periodically with the weights of the value network. In between updates this provides the agent with fixed EFE-value or Q-value targets, such that the value network does not have to chase a constantly moving objective.

The VAE of the POMDP dAIF agent is pre-trained to deconstruct input images into a distribution over latent states and to subsequently reconstruct them as accurately as possible.
\begin{figure}[hbtp!]
\includegraphics[width=\textwidth]{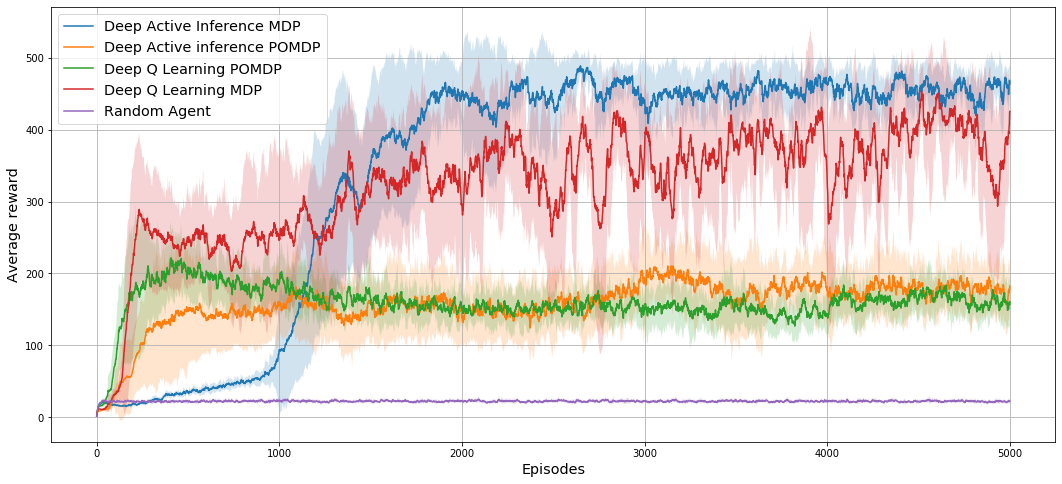}
\caption{Average reward comparison for the CartPole-v1 problem.}
\label{fig:results}
\end{figure}

Fig. \ref{fig:results} shows the mean and standard deviation of the moving average reward ($MAR$) over all runs for the five algorithms at each episode. Each agent performed 10 runs of 5000 episodes. The moving average reward for an episode $e$ is calculated using an smoothing average:
\begin{align}
    MAR_e = 0.1CR_e + 0.9MAR_{e-1}
\end{align}
Where $CR_e$ is the cumulative reward of episode $e$ and $MAR_{e-1}$ is the $MAR$ of the previous episode. 

The dAIF MDP agent results closely resemble those presented in \cite{MILLIDGE2020102348} and outperforms the DQN MDP agent by a significant margin. Further, the standard deviation shadings show that the dAIF MDP is agent is more consistent between runs than the DQN agent. The POMDP agents are both demonstrated to be capable of learning successful policies, attaining comparable performance.

We have exploited probabilistic model based control through a VAE that encodes the state. On the one hand, this allows the tracking of an internal state which can be used for a range of purposes, such the planning of rewarding policies and the forming of expectations about the future. On the other hand, it makes every part of the algorithm dependent on the proper encoding of the latent space, conversely to the DQN that did not require a state representation to achieve the same performance. However, we expect our approach to improve relative to DQN in more complex environments where the world state encoding can play a more relevant role.

\section{Conclusion}
We described a dAIF model that tackles partially observable state problems, i.e., it learns the policy from high-dimensional inputs, such as images. Results show that in the MDP case the dAIF agent outperforms the DQN agent, and performs more consistently between runs. Both agents were also shown to be capable of learning (less) successful policies in the POMDP version, where the performance between dAIF and DQN models was found to be comparable. Further work will focus on validating the model on a broader range of more complex problems.

\section*{Appendix}
\definecolor{c0}{HTML}{A5B9DA}
\definecolor{c1}{HTML}{CADAEF}
\definecolor{c2}{HTML}{D2E5F3}
\definecolor{c3}{HTML}{FFFFFF}
\setlength\arrayrulewidth{1pt}

\begin{center}
\def\arraystretch{1.3}
{\rowcolors{2}{c2}{c3}
\begin{tabular}{ |p{3.5cm}|p{8.5cm}|  }
 \hline
 \rowcolor{c0}
 \multicolumn{2}{|c|}{\textbf{Deep Q Agent MDP}} \\
 \hline
 \rowcolor{c1}
 \multicolumn{1}{|c|}{Networks \& params.}   & \multicolumn{1}{c|}{Description} \\
 $N_s$ & Number of states. \\
 $N_a$ & Number of actions. \\
 Q-value network & Fully connected network using an Adam optimizer with a learning rate of $10^{-3}$, of the form: $N_s \times 64 \times N_a$. \\
 $\gamma$ & Discount factor set to 0.98 \\
 Memory size & Maximum amount of transition that can be stored in the memory buffer: 65,536 \\
 Mini-batch size & 32 \\
 Target network freeze period & The amount of time steps the target network’s parameters are frozen, until they are updated with the parameters of the value network: 25 \\
 \hline
\end{tabular}
}
\end{center}

\begin{center}
\def\arraystretch{1.3}
{\rowcolors{2}{c2}{c3}
\begin{tabular}{ |p{3.5cm}|p{8.5cm}|  }
 \hline
 \rowcolor{c0}
 \multicolumn{2}{|c|}{\textbf{Deep Q Agent POMDP}} \\
 \hline
 \rowcolor{c1}
 \multicolumn{1}{|c|}{Networks \& params.}   & \multicolumn{1}{c|}{Description} \\
 $N_a$ & Number of actions. \\
 Q-value network & Consists of three 3D convolutional layers (each followed by batch normalization and a rectified linear unit) with $5 \times 5 \times 1$ kernels and $2 \times 2 \times 1$ strides with respectively 3, 16 and 32 input channels, ending with 32 output channels. The output is fed to a $2048 \times 1024$ fully connected layer which leads to a $1024 \times N_a$ fully connected layer. Uses an Adam optimizer with the learning rate set to $10^{-5}$. \\
 $\gamma$ & Discount factor set to 0.99 \\
 Memory size & Maximum amount of transition that can be stored in the memory buffer: 65,536 \\
 Mini-batch size & 32 \\
 Target network freeze period & The amount of time steps the target network’s parameters are frozen, until they are updated with the parameters of the value network: 25 \\
 \hline
\end{tabular}
}
\end{center}

\begin{center}
\def\arraystretch{1.3}
{\rowcolors{2}{c2}{c3}
\begin{tabular}{ |p{3.5cm}|p{8.5cm}|  }
 \hline
 \rowcolor{c0}
 \multicolumn{2}{|c|}{\textbf{Deep Active Inference Agent MDP}} \\
 \hline
 \rowcolor{c1}
 \multicolumn{1}{|c|}{Networks \& params.}   & \multicolumn{1}{c|}{Description} \\
 $N_s$ & Number of states. \\
 $N_a$ & Number of actions. \\
 Transition network & Fully connected network using an Adam optimizer with a learning rate of $10^{-3}$, of the form: $(N_s+1) \times 64 \times N_s$. \\
 Policy network & Fully connected network using an Adam optimizer with a learning rate of $10^{-3}$, of the form: $N_s \times 64 \times N_a$, a softmax function is applied to the output. \\
 EFE-value network & Fully connected network using an Adam optimizer with a learning rate of $10^{-4}$, of the form: $N_s \times 64 \times N_a$. \\
 $\gamma$ & Precision parameter set to 1.0 \\
 $\beta$ & Discount factor set to 0.99 \\
 Memory size & Maximum amount of transition that can be stored in the memory buffer: 65,536 \\
 Mini-batch size & 32 \\
 Target network freeze period & The amount of time steps the target network’s parameters are frozen, until they are updated with the parameters of the value network: 25 \\
 \hline
\end{tabular}
}
\end{center}

\begin{center}
\def\arraystretch{1.3}
{\rowcolors{2}{c2}{c3}
\begin{tabular}{ |p{3.5cm}|p{8.5cm}|  }
 \hline
 \rowcolor{c0}
 \multicolumn{2}{|c|}{\textbf{Deep Active Inference Agent POMDP}} \\
 \hline
 \rowcolor{c1}
 \multicolumn{1}{|c|}{Networks \& params.}   & \multicolumn{1}{c|}{Description} \\
 \raggedright $N_l$ & Size of the VAE latent space, here set to 32. \\
 $N_a$ & Number of actions. \\
 Encoder-network $q_\theta(s_t|o_{t-3:t})$ & Consists of three 3D convolutional layers (each followed by batch normalization and a rectified linear unit) with $5 \times 5 \times 1$ kernels and $2 \times 2 \times 1$ strides with respectively 3, 16 and 32 input channels, ending with 32 output channels. The output is fed to a $2048 \times 1024$ fully connected layer which splits to two additional $1024 \times N_l$ fully connected layers. Uses an Adam optimizer with the learning rate set to $10^{-5}$. \\
 Decoder-network $p_\vartheta(o_{t-3:t}|z_t)$ & Consists of a $N_l \times 1024$ fully connected layer leading to a $1024 \times 2048$ fully connected layer leading to three 3D transposed convolutional layers (each followed by batch normalization and a rectified linear unit) with $5 \times 5 \times 1$ kernels and $2 \times 2 \times 1$ strides with respectively 32, 16 and 3 input channels, ending with 3 output channels. Uses an Adam optimizer with the learning rate set to $10^{-5}$. \\
 Transition-network $f_\phi(s_{\mu,t}, a_{t})$ & Fully connected network using an Adam optimizer with a learning rate of $10^{-3}$, of the form: $(2 N_l+1) \times 64 \times N_l$. \\
 Policy-network $q_\xi(s_{\mu,t},s_{\Sigma,t})$ & Fully connected network using an Adam optimizer with a learning rate of $10^{-3}$, of the form: $2 N_l \times 64 \times N_a$, a softmax function is applied to the output. \\
 EFE-value-network $f_\psi(s_{\mu,t},s_{\Sigma,t})$ & Fully connected network using an Adam optimizer with a learning rate of $10^{-4}$, of the form: $2 N_l \times 64 \times N_a$. \\
 $\gamma$ & Precision parameter set to 12.0 \\
 $\beta$ & Discount factor set to 0.99 \\
 $\alpha$ & A constant that is multiplied with the VAE loss to take it to the same scale as the rest of the VFE terms, set to $4 \times 10^{-5}$ \\
 Memory size & Maximum amount of transition that can be stored in the memory buffer: 65,536 \\
 Mini-batch size & 32 \\
 Target network freeze period & The amount of time steps the target network’s parameters are frozen, until they are updated with the parameters of the value network: 25 \\
 \hline
\end{tabular}
}
\end{center}

\bibliographystyle{splncs04}
\bibliography{references}

\end{document}